# Anomaly Detection and Improvement of Clusters using Enhanced K-Means Algorithm

Vardhan Shorewala
*University of California, Berkeley*
vardhanshorewala@berkeley.edu

Shivam Shorewala
*University of California, Berkeley*
shivamshorewala@berkeley.edu

*Abstract* — **This paper explores a unified approach to the improvement of clusters and the detection of anomalies in a dataset. It presents a novel approach for the formation of tighter and better clusters than traditional methods such as the K-means algorithm. It is evaluated against intrinsic measures for unsupervised learning such as the silhouette coefficient, Calinski Harabasz index, and the Davies Bouldin index. The proposed method decreases the intra-cluster variance of N clusters, until the variance approaches a global minimum. The method also extends K-means as a novel method for the detection of anomalies in the dataset. The proposed algorithm's performance is evaluated by extrinsic measures such as the Jaccard similarity score, the V-measure cluster, and the F1 Score. The algorithm is tested upon synthetic and real datasets, UCI Breast Cancer and UCI Wine Quality, to showcase its effectiveness in both cases. The proposed algorithm reduced the variance of the synthetic and real dataset, Wine Quality, by 18.7% and 88.1% respectively. It was also effective in increasing the accuracy and F1 score by 22.5% and 20.8% in the case of the Wine Quality dataset.**

I.  INTRODUCTION

Outlier detection is an important task in data analysis. An outlier is defined as a point that significantly varies from the rest [1]. Outliers can arise from human error, faults in the system, instrument errors, or from malicious activity [2]. Detection of outliers has two major applications. Firstly, it helps in noise detection in datasets. The presence of unwanted noise can lead to the formation of an incorrect model, producing a skewed output [3]. Therefore, the effective removal of noise is an important step in the processing of raw data. Secondly, outlier detection has industry applications, where detecting sensitive outliers is the task of the algorithm itself. This includes fraud detection, medical datasets, etc.

Despite their close complementarity, clustering and anomaly detection are often considered as separate issues. However, there is an inherent number of outliers present in the clusters formed. The K-means algorithm is sensitive to outliers, causing the centroids to adjust accordingly. This could have a disproportionate effect upon the classification with many false negatives present in the result. This means that the practicality of the model reduces due to adjusted positions of the centroids in relation to outliers in the clusters.

A common approach to solve this is to remove data-points directly from the dataset, using methods such as the Z-value test or standard deviation. However, outliers may not always be present towards the skewed ends of a probability distribution graph of the data and may be prominent near the head of the curve itself. Therefore, detection of anomalies using clusters does not limit itself to the tails of a probability distribution graph and, instead, makes use of the density of data points found in a certain region. This allows for the effective removal of anomalies and outliers through the entirety of the dataset, including local outliers. A robust method of K-means will allow for the detection of anomalies, alongside, improving the quality of clusters. This will allow for the removal of local outliers as well rather than being limited to global outliers as by current methods. The proposed method is shown in figure 1.

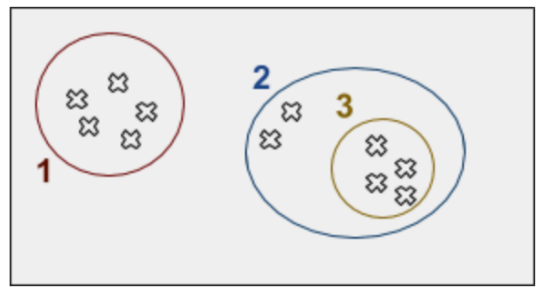

Fig. 1. The image depicts how K-means clusters are affected by anomalies. The removal of anomalies is clearly depicted progression of cluster (2) to cluster (3)

Figure 1 depicts how the clustering algorithm will find anomalies, which would not have been found using existing methods. The algorithm is based on the density of points found around the centroid of the cluster. Figure 1 also clearly depicts that the progression of cluster (2) to cluster (3) via the proposed algorithm will allow for the creation of tighter clusters, which will not produce false negatives in the output.

The algorithm will be evaluated against intrinsic measures such as the silhouette coefficient and the average cluster variance, which showcase its effectiveness in unsupervised learning with no ground truth labels. However, the algorithm would be further evaluated against extrinsic measures such as the Jaccard similarity score to depict its effectiveness in supervised learning with ground truth labels as well. The algorithm will be tested on both real and synthetic datasets to showcase its effectiveness in both circumstances.

II. RELATED WORK

A lot of the current research is directed towards finding global outliers in a dataset i.e., outliers found towards the tails of a probability distribution graph. However, there are much fewer methods focusing upon the removal of local outliers present in the dataset, which can hamper with clustering models. Furthermore, many techniques such as density-based spatial clustering of applications (DBSCAN) [4] have been studied extensively; however, only few of the studies are based on K-means, and, therefore, this study will be focusing upon the K-means algorithm.

Most of the previous studies on outlier detection were performed in statistics. These studies can be sorted into two major categories. The first is distribution-based in which the data is modelled using a standard distribution model such as Gaussian and plotted against a probability graph. Several tests have been developed in this category known as discordancy tests for different scenarios [5]. However, a prominent drawback is that most of these distributions are univariate i.e., they are unable to model multivariate data, which is important in large datasets with multiple features that are commonly used now. Furthermore, for knowledge discovery in database (KDD) applications, most of their distributions remain unknown. Therefore, fitting multivariate data to these distributions will not produce proper outputs. The second category to detect outliers in the field of statistics is depth-based. Each data object is represented as a point in a k-dimensional space and is assigned a depth. Outliers are more likely to be data objects with smaller depths [6, 7]. Depth-based outlier-detection algorithms work well for $k = 2\ and\ k = 3$ [8, 9], but become inefficient for $k \geq 4$. This is because they rely on the computation of $k - d$ convex hulls, which have a lower bound complexity of $\Omega\left(n^{\frac{k}{2}}\right)$, $n$ being the number of objects.

Knorr and Raymond put forward a distance-based method to find outliers in the data. They proposed methods of finding outliers in both gaussian and non-gaussian distributions. Knorr, Raymond and Tucakov generalized their methods for $K-$dimensional datasets [10]. This had a complexity of $O\left(kN^2\right)$, where $k$ is the dimensionality and $N$ is the number of objects. They provide a more efficient method to compute the top $n$ outliers in higher dimensionality data as compared to the depth-based approaches. However, the methods proposed are limited to global outliers, and do not include finding local outliers.

Even though research has been done in finding local outliers, few algorithms perform clustering and outlier detection at the same time. The K-means with outlier removal (KMOR) algorithm developed by Gan and Kwok-Po partitions the dataset into $k + 1$ clusters, which include $k$ clusters and a group of outliers that cannot fit into the $k$ clusters [11]. The KMOR algorithm assigns all outliers into a group naturally during the clustering process. Ahmed and Naser created the ODC (outlier detection and clustering) algorithm which aids in outlier detection as well. In the ODC algorithm, a point is considered an outlier if it is $p$ times the average distance from the centroid of the cluster it is found in [12]. Chawla and Gionis also put forth an algorithm to perform outlier detection and clustering at the same time, which requires the parameters $k\ and\ l$: the number of desired clusters and the desired number of outliers to be found [13].

Hautamäki et al. proposed the ORC (Outlier Removal Clustering) algorithm [14]. The method employs both clustering and outlier discovery to improve estimation of the centroids of the generative distribution. There are two steps to the algorithm: the first consists purely of K-means while the second stage is an iterative process that removes vectors that are far from the centroids of the respective cluster. Schelling and Plant have also proposed the K-Means with Noise (KMN) algorithm which helps reduce noise in the dataset [15]. The algorithm finds outliers in the area bound by the intersection of hyperplanes; the hyperplane is found between two neighboring Voronoi cells. Beer et al. developed the k-Means based Outlier Removal algorithm (MORe++) which is aimed to detected outliers in high dimensional data [16]. The algorithm looks at the data-points in each cluster separately. One-dimensional data projections become meaningful and finding one-dimensional outliers becomes easier. Franti and Yang created an algorithm to reduce noise K-means by processing each point in dataset and then replacing it by the medoid of the neighborhood. This processing could be iterated to help reduce the noise of the dataset [17]. Chen, Wang, et al. created the Multi-View Clustering with Outlier Removal (MVCOR) algorithm, which is aimed at reducing noise in multi-view clustering. The algorithm defines two forms of outliers and sorts the outliers into two clusters, cleaning the dataset in the process. This has shown to improve multi-view clustering methods, which are sensitive to outliers [18]. Lastly, Xie, Zhang, Lim, et al. proposed two variants of Firefly Algorithm, an improved K-means clustering algorithm. It resolved the issues related to the initialization sensitivity and the local optima traps of the K-means algorithm [19].

III. PROPOSED ALGORITHM

This section introduces the algorithm proposed, and the mathematical notation for the architecture of the design.

We consider a set of $z$ data-points in $k - $ dimensional space, given by $X = \{x_1, x_2 \ldots x_z\}$. We then consider a set of $c$ centroids found by the classical K-mean algorithm. The set of clusters is denoted by $C = \{c_1, c_2 \ldots c_n\}$, where $n$ is the number of desired clusters. We then assign each $x$ ($x \subseteq X$) to the nearest $c$ ($c \subseteq C$). To do this, we first calculate the $k$-dimensional Euclidean distance of each point in a cluster from the respective centroid. The general formula is shown in equation one. Euclidean distance was used as the algorithm only supports Minkowski style metrics currently.

$$d(x_i, y_i) = \left( \sum_{i=1}^{k} (|| x_i - y_i ||)^2 \right)^{\frac{1}{2}} \quad (1)$$

This general distance function $d(x_i, y_i)$ is applied for finding the distance of a point to each of the $n$ centroids ($c$) in the following manner:

$$d_n = \left[ d(x_i, c_i) \right]_1^n \quad (2)$$

Then we compute $argmin(d_n)$ to find the closest centroid for a point and then assign it to that respective cluster. This process is iterated until each $k$-dimensional point has been allotted to the nearest $c$ ($c \subseteq C$).

Once each point has been allotted to a cluster, the average variance, $V$, is calculated across a cluster with $j$ points and centroid $c$:

$$V^2 = \frac{\sum_{i=1}^{j} (x_i - c)^2}{j - 1} \quad (3)$$

The variance is calculated across each of the $n$ clusters and averaged:

$$V_{avg} = \frac{\sum_{i=1}^{n} V_i}{n} \quad (4)$$

Once the initial average variance is calculated, the algorithm begins outlier removal. For a given cluster with centroid $c$, the euclidean distance from the centroid is calculated to each point of the $j$ points in that cluster.

$$d_c = \left[ d(x_i, c) \right]_{i=1}^{j} \quad (5)$$

After receiving an output of all the $k-$ dimensional euclidean distances, all points greater than 2 standard deviations from the mean are labelled as outliers and removed from the respective cluster. The value of 2 can be changed according to user input; however, 2 was calculated mathematically. The array of Euclidean distance ($d_c$) was found not to follow a gaussian distribution. Therefore, the probability of points lying in a certain region was estimated using Chebyshev's inequality. The minimum number of points classified as non-outliers was set to 75%.

$$P[|k - \mu| \geq k\sigma] \geq 1 - \frac{1}{k^2} \quad (6)$$

$$= P[|k - \mu| \geq 2\sigma] \geq 0.75$$

This distance-based method of removing outliers is conducted for each of the $n$ clusters.

Once the outliers are removed, the centroids are re-positioned to the modified dataset, and the outliers are then removed from that modified dataset using the same algorithm depicted above. The lower limit for Chebyshev's inequality remains at a constant of 2. This process is iterated until the variance converged to a global minimum. The condition is shown below:

$$\lim_{a \to 0} : a = \frac{\Delta}{\Delta V} [V_n, V_{n-1}] \quad (7)$$

**Algorithm 1** Enhanced K-Means

**Input:** A set of points $X$, $X = \{x_1, x_2 ... x_z\}$
**Output:** A set of outliers Y, $Y = \{y_1, y_2 ... y_i\}$
**Functions:** Standard K-Means to compute a set of $n$ centroids, $C$: $C = \{c_1, c_2 ... c_n\}$
A $k$-dimensional Euclidean distance, denoted by $d(x_i, y_i)$
Average intra-cluster variance, denoted by $V'$

1: **while** $V'$ does not converge **do**
2:    compute $C$: $C = \{c_1, c_2 ... c_n\}$
3:    **for** $x$ in $X$ **do**
4:      $argmin(d(x, c_1^n))$
5:    let $D_1^n$ be sets of all points in clusters $c_1^n$
6:    **for** $v$ in range (length ($C$)) **do**
7:      $t_v$ = mean ($D_v$) + $2\sigma$
8:      **for** $p$ in $D_v$ **do**
9:        **if** $d(p, c_v) \geq t_v$ **do**
10:          delete $p$ from $D_v$
11:          add $p$ to $Y$
12: **return** $Y$, $Y = \{y_1, y_2 ... y_i\}$

IV.               EXPERIMENTAL SETUP

The algorithm was tested upon two datasets: synthetic and real data. The enhanced K-means was evaluated using intrinsic and extrinsic measures; however, emphasis was provided to intrinsic as the algorithm aimed to improve unsupervised clustering. Therefore, the synthetic dataset had no ground truths and was evaluated solely on intrinsic measures while the real data had ground truth labels and was evaluated on both extrinsic and intrinsic measures. Intrinsic measures generally evaluate clusters upon their separation and compactness. In addition to the intra-cluster variance used in the algorithm, three other intrinsic measures were used to evaluate the clusters

Firstly, the Silhouette index ($S$) [20] can be used to find the separation distance between the clusters that are formed. It has a complexity of $O(n^2)$ and is denoted by:

$$S = \frac{b - a}{\max(a, b)} \quad (8)$$

*where a is the mean intra-cluster distance and b is the mean nearest-cluster distance*

Secondly, the Calinski Harabasz Index [21], Variance Ratio Criterion, is the ratio between the within cluster dispersion and the between-cluster dispersion.

*For a data E with size $n_e$ and k clusters, the index is:*

$$c = \frac{tr(B_k)}{tr(W_k)} \times \frac{n_e - k}{k - 1} \quad (9)$$

where $tr(B_k)$ is the trace of the between group dispersion matrix and $tr(W_k)$ is the trace of the within-cluster dispersion matrix.

Thirdly, the Davies-Bouldin Index [22] index signifies the average similarity measure of each cluster with the cluster most similar to it. A better score indicates clusters are further apart and less dispersed.

There were two extrinsic measures used for evaluation of the model with ground truth labels in addition to accuracy and F1 score. Firstly, the Jaccard Coefficient is defined as the intersection over union of two sets i.e., actual labels against the predicted labels. It is a statistic used for gauging the similarity and diversity of sample sets.

$$J(A, B) = \frac{|A \cap B|}{|A \cup B|} \quad (10)$$

Secondly, the V-measure was used, which is the harmonic mean between homogeneity and completeness of the true ground truth labels against the predicted labels, and is defined by:

$$v = \frac{(1 + \beta) \times homogeneity \times completeness}{\beta + homogeneity + completeness} \quad (11)$$

V. SYNTHETIC DATA

A 2-dimensional dataset was generated to allow for easier data visualization, consisting of a thousand randomly generated points. These points had no ground truth labels and were evaluated solely based on unsupervised learning algorithms. A default of 5 centroids were used and explored in detail using the enhanced K-means algorithm. However, the performance of the algorithm is evaluated against a varying number of centroids as well. Figure 2 evaluates the algorithm for the synthetic dataset based upon intrinsic measures for 5 clusters. It is plotted against the iterations to showcase how to metrics vary with each loop of outlier removal.

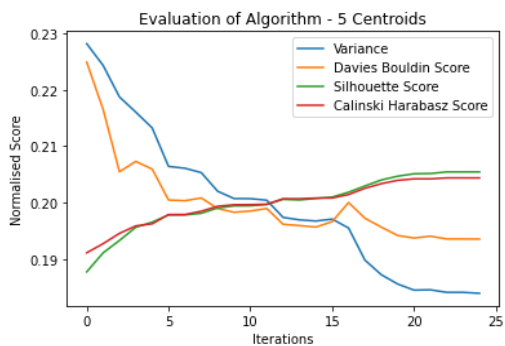

Fig. 2. The image shows the evaluation of the model based on normalized values of metrics for 5 clusters.

Figure 2 clearly depicts the algorithm's effectiveness upon the synthetic dataset. The number of points that were classified as anomalies were 75, which represents 7.5% of the dataset. However, this value can be adjusted by modifying the bound for Chebyshev's inequality. It also depicts how the variance clearly approaches a global minimum and reduces by around 18.7%.

The other metrics used for unsupervised learning were normalized to represent the trend easily. The Davies Boldin Score decreased by 13.9% as compared to traditional K-means, depicting the algorithm's effectiveness in decreasing the ratio between the cluster scatter and separation. On the other hand, the Calinski Harabasz Score and Silhouette Score increased by 6.95% and 9.44% respectively in comparison to traditional K-means. This showcases that clusters are now tighter and more well-defined. The classification of points by the algorithm are shown in figure 3.

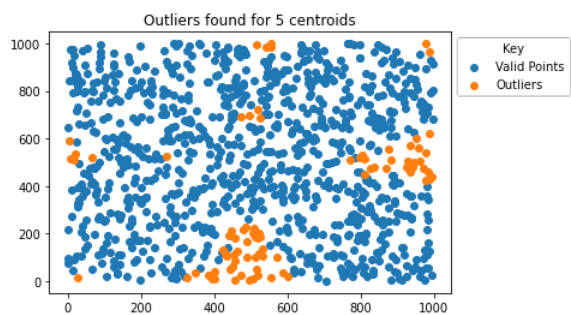

Fig. 3. The image shows the local and global outliers detected by the algorithm for 5 clusters.

Figure 3 clearly depicts the algorithm's effectiveness in classifying both local and global points as outliers. The percentage of outliers can be adjusted by modifying Chebyshev's inequality.

Figure 4 depict the algorithm's effectiveness on a synthetic dataset in a 2-dimensional area with the number of centroids ranging from 2 to 10. The graph shows the absolute, overall percentage change for each unsupervised metric for the respective number of clusters. The metrics are averages for ten randomly generated synthetic datasets to increase consistency of results. This is to evaluate the performance of the algorithm for larger number of clusters, which are tighter.

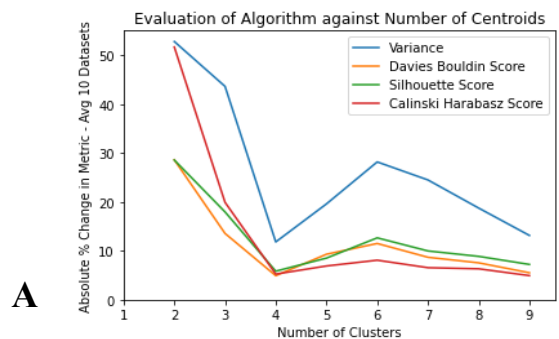

A

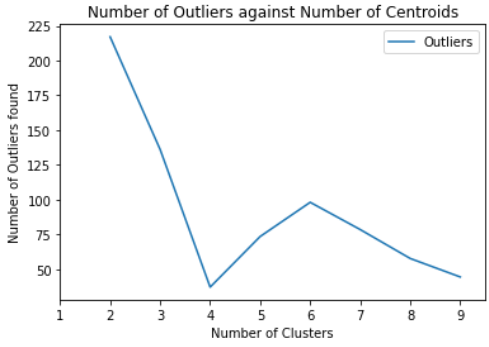
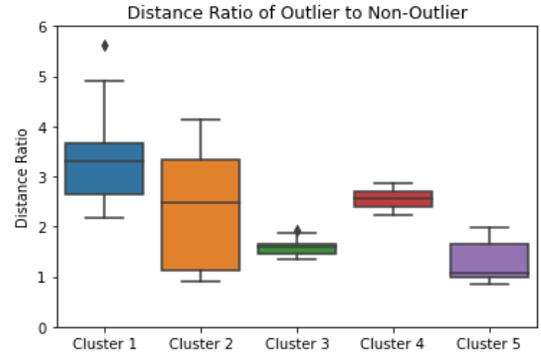

Fig. 4. Evaluation of the algorithm for a varying number of clusters. **A** shows the absolute percentage change in metric for the respective number of clusters. **B** shows the outliers found for the respective number of clusters.

Fig. 5. Box plot showing the ratio of the distance of an outlier to the average distance to a non-outlier in a respective cluster.

Figure 4 shows how the algorithm functions on larger numbers of centroids as well. However, the effectiveness of algorithms reduces. This is because the clusters become tighter as the number of centroids increase, reducing the number of outliers that deviate from the mean distance. The uniform distribution of points, as shown in figure 3, helps reduce the effectiveness of the algorithm for larger numbers of centroids as well. However, the algorithm was effective in reducing the intra-cluster variance by 13% for nine centroids, as shown in figure 4.

Figure 5 clearly depicts how the distance to an outlier is much greater to that of a non-outlier. This varies for each cluster, with cluster 1 having the highest, an average of 3.32, while cluster 5 has the least, an average of 1.33. Figure 6 depicts the averaged metric of ten iterations for the algorithm on the WBC dataset.

## VI. REAL DATA

The proposed algorithm was tested upon the Breast Cancer Wisconsin (WBC) dataset, which was taken from UCI [23]. A similar analysis was performed as with synthetic data; however, the algorithm was also evaluated based on supervised learning techniques here. The dataset contained 31 numerical attributes and one categorical which can be interpreted as the class label. This results in the formation of a 31-dimensional space. The data set contained 569 instances, and two distinct class labels. We set the number of clusters as 5 and continue to use the default 75% in Chebyshev's inequality. The analysis performed is similar to that with the synthetic dataset; however, it includes metric evaluating the effectiveness of the algorithm in supervised learning as well.

The ratio of the distance to an outlier to the average non-outliers was calculated for each cluster. This was calculated using the following formula, where $o_i$ represents an outlier, c is the centroid, and $x_i$ represents the non-outliers in that cluster. The total non-outliers are n and outliers are z.

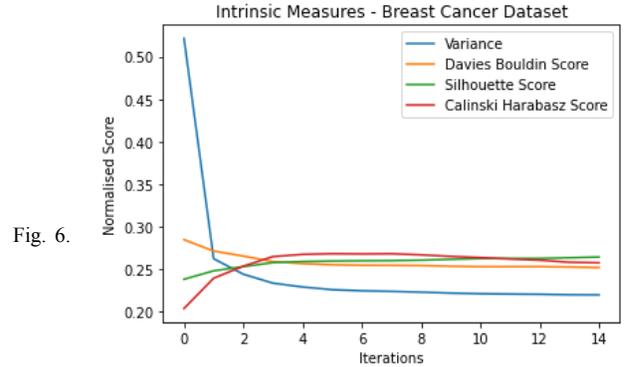

Fig. 6.

$$\text{Distance Ratio} = \frac{\left[d(o_i, c)\right]_{i=1}^{z}}{\frac{\sum_{i=1}^{n} d(x_i, c)}{n}} \quad (12)$$

The distance ratio was plotted against the cluster number in the form of a box plot. This will allow us to visualize the spread of the outliers in each cluster. The total number of outliers was found to be 58 using the proposed algorithm, which is 10.1% of the dataset. This distance ratio of the outlier to non-outlier of each cluster is shown in figure 5.

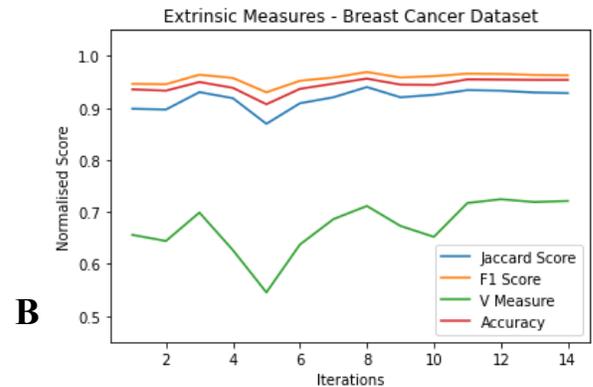

Evaluation of algorithm on WBC dataset, with number of centroids set as 5. **A** shows the intrinsic measures. **B** shows the extrinsic measures.

Figure A clearly shows the efficiency and effectiveness of the algorithm in unsupervised learning whilst using a real dataset. The intra-cluster variance and Boldmin index reduced by 57.9% and 11.5% respectively as compared to traditional K-means. This clearly shows how the new formed clusters are tighter and more distinct as compared to original ones. Furthermore, the Silhouette Score and Harabasz Score increased by 11.0% and 31.7% respectively in comparison to regular K-means. This helps show how the clusters are more uniform and well-defined in the 31-dimensions.

Figure B shows the performance of the algorithm in supervised learning in K-means. It depicts how the algorithm can improve each metric used to evaluate the clusters with ground truth labels. Although the improvement is small, the algorithm always improves or maintains the same level of the metric as the initial. This showcases the effectiveness of the algorithm for both supervised and unsupervised cases. The accuracy and F1 metric increased by 1.95% and 1.73% respectively. This means the number of correct classifications produced by the algorithm is higher. It also has lower false positive rate and false negative rates. Furthermore, the Jaccard Score and V Measure increased by 3.34% and 9.95% respectively as compared to traditional K-means. This means the predicted labels and the ground truths had greater similarity. Furthermore, the prediction approached symmetry as the following condition was neared:

$$homogeneity(b, a) = completeness(a, b) \qquad (13)$$

The proposed algorithm was also tested upon the Red Wine Quality dataset, taken from UCI [24]. A similar study to the WBC dataset was conducted. The dataset contained 11 numerical attributes and 1600 instances. However, the target variable, the quality of wine, was multi-label i.e., non-binary. This was to evaluate the performance of the algorithm for multi-label target classes in supervised learning. We set the number of clusters as 5 and continue to use the default 75% in Chebyshev's inequality. Figure 7 shows the extrinsic and intrinsic measures for the Red Wine Quality dataset.

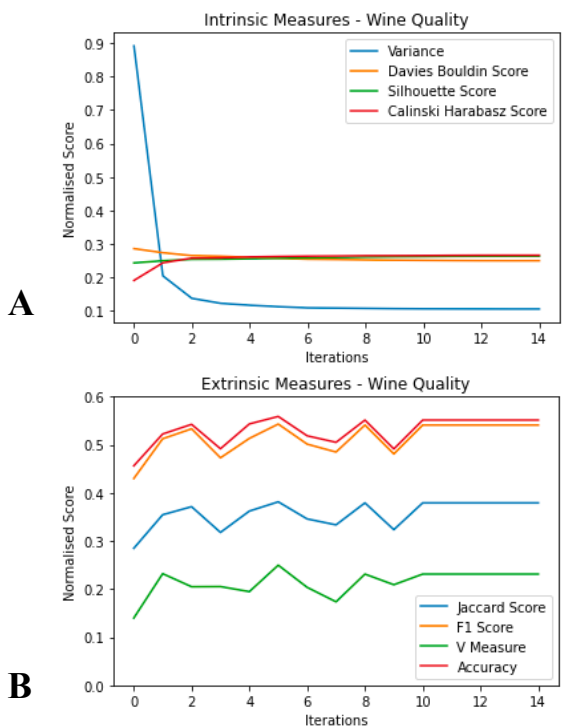

Fig. 7. Evaluation of algorithm on Red Wine Quality dataset, with number of centroids set as 5. **A** shows the intrinsic measures. **B** shows the extrinsic measures.

Figure 7 clearly show the algorithm's effectiveness in datasets containing multi-label target classes. The algorithm found 121 outliers in the dataset, which is 7.6% of the dataset. Figure A shows the case of unsupervised learning. The intra-cluster variance and Boldmin index reduced by 88.1% and 12.6% respectively as compared to traditional K-means. This showcases the large improvement to the traditional clusters formed. Additionally, the clusters became more symmetrical as the Silhouette Score and Harabasz Score increased by 8.2% and 39.4% respectively. Figure B showcases the effectiveness in the case of supervised metrics. The weighted average was used for the Jaccard and V-Measure score as they were multi-labels. The algorithm was able to increase the correct classifications made by the algorithm as the accuracy and F1 metric increased by 22.5% and 20.8% respectively. This showcases the effectiveness of the algorithm in cases of supervised learning. Furthermore, the Jaccard Score and V Measure increased by 22.5% and 78.6% respectively as compared to traditional K-means. This depicts how the proposed algorithm reduces the false positives and false negatives by a large percentage.

VII. DISCUSSION

Clustering and outlier detection has been treated as separate problems; however, they are complementary. Local outliers can skew the position of the centroids, while cluster shape and spread can indicate possible outliers in the data.

A unified approach is showcased and depicts an enhanced version of the K-means algorithm. Through an iterative process, the algorithm depicts how the intra-cluster variances approach a global minimum. Furthermore, the number of outliers removed can be modified by changing the bounds of Chebyshev's inequality. The algorithm was evaluated based on both intrinsic and extrinsic techniques. The positive results showed the efficiency and usefulness in both cases of clustering: supervised and unsupervised.

The proposed algorithm was tested upon synthetic data first, without ground truth labels. This confirmed that the algorithm works efficiently in a controlled environment with a regular spread of data. This algorithm was then tested upon real datasets, UCI-licensed Wisconsin Breast Cancer and Red Wine Quality Dataset. The algorithm halved the intra-cluster variance, whilst only removing 10.1% of points in the case of the WBC dataset. Furthermore, the dataset had ground truth labels. The algorithm showed high performance in the case of supervised learning here as well, bettering the extrinsic measures used for evaluation of the algorithm.

This algorithm can be used for noise detection in datasets which require cleaning. The algorithm improves upon K-means, which uses centroids to create spherical clusters to group points. This means the algorithm will perform best on a dataset with a dataset with a Von Mises-Fisher distribution.

However, as shown by the WBC dataset, it performs well upon normal distributions as well. Filtering local noise aids in the improvement of clustering algorithms, which is sensitive to noise. Furthermore, this algorithm can be used to identify local and global outliers in datasets, which is shown in the medical dataset it was tested upon. Additionally, this has applications in fraud detection, network anomaly detection, etc.

Future work involves automating the process of choosing the bound of Chebyshev's inequality. This will create a dynamic bound for the inequality, specific to the spread of data in a certain dataset. Furthermore, the algorithm only supports Minkowski style metrics currently. Future work could extend the algorithm to allow for other metrics such as normalized spatial, angular, etc., based on user input.